\documentclass[a4paper,conference]{IEEEtran}
\usepackage{times,xcolor}
\usepackage{footnote}
\usepackage{latexsym}
\usepackage{afterpage}
\usepackage[utf8]{inputenc}
\usepackage{graphicx}
\usepackage{subfig}
\usepackage{mathrsfs}
\usepackage{array,tabularx}
\usepackage{amsmath, stmaryrd, amssymb}
\usepackage[stable]{footmisc} 
\usepackage{bm,mathtools,upgreek}
\usepackage{multirow}
\usepackage{subfiles}
\usepackage{comment}
\usepackage[pdftex,
            pdfauthor={Ismail Oussaid, William Vanhuffel, Pirashanth Ratnamogan, Mhamed Hajaiej, Alexis Mathey, Thomas Gilles},
            pdftitle={Information Extraction from Visually Rich Documents with Font Style Embeddings},
            pdfsubject={Information extraction (IE) from documents is an intensive area of research with a large set of industrial applications. Current state-of-the-art methods focus on scanned documents with approaches combining computer vision, natural language processing and layout representation. We propose to challenge the usage of computer vision in the case where both token style and visual representation are available (i.e native PDF documents). Our experiments on three real-world complex datasets demonstrate that using token style attributes based embedding instead of a raw visual embedding in LayoutLM model is beneficial. Depending on the dataset, such an embedding yields an improvement of 0.18\% to 2.29\% in the weighted F1-score with a decrease of 30.7\% in the final number of trainable parameters of the model, leading to an improvement in both efficiency and effectiveness. },
            pdfkeywords={Information Retrieval; Natural Language Processing;Computer Vision and Pattern Recognition;Machine Learning;LayoutLM}]{hyperref}
            
\renewcommand{\vec}[1]{\bm{\mathrm{#1}}}
\usepackage{microtype}

\usepackage{pgf-pie}  

\usepackage[english]{babel}

\usepackage{hyperref}
\usepackage{mathtools}          

\newenvironment{conditions*}
  {\par\vspace{\abovedisplayskip}\noindent
   \begin{tabular}{>{$}l<{$} @{} >{${}}c<{{}$} @{} l}}
  {\end{tabular}\par\vspace{\belowdisplayskip}}

\newcommand\BibTeX{B\textsc{ib}\TeX}
\def\BibTeX{{\rm B\kern-.05em{\sc i\kern-.025em b}\kern-.08em
    T\kern-.1667em\lower.7ex\hbox{E}\kern-.125emX}}

\title{Information Extraction from Visually Rich Documents with Font Style Embeddings
}

\author{ 
\IEEEauthorblockN{Ismail Oussaid}
\IEEEauthorblockA{{BNP Paribas} \\
ismail.oussaid@student-cs.fr}
\and
\IEEEauthorblockN{William Vanhuffel}
\IEEEauthorblockA{{BNP Paribas} \\
william.vanhuffel@bnpparibas.com \\
}
\and
\IEEEauthorblockN{Pirashanth Ratnamogan}
\IEEEauthorblockA{{BNP Paribas} \\
pirashanth.ratnamogan@bnpparibas.com}
\and
\IEEEauthorblockN{Mhamed Hajaiej}
\IEEEauthorblockA{{BNP Paribas} \\
mhamed.hajaiej@bnpparibas.com}
\and
\IEEEauthorblockN{Alexis Mathey}
\IEEEauthorblockA{{BNP Paribas} \\
alexis.mathey@bnpparibas.com}
\and
\IEEEauthorblockN{Thomas Gilles}
\IEEEauthorblockA{{BNP Paribas} \\
thomas.gilles@bnpparibas.com}
}

\date{}

\begin{document}
\maketitle
\begin{abstract}
Information extraction (IE) from documents is an intensive area of research with a large set of industrial applications. Current state-of-the-art methods focus on scanned documents with approaches combining computer vision, natural language processing and layout representation. We propose to challenge the usage of computer vision in the case where both token style and visual representation are available (i.e native PDF documents). Our experiments on three real-world complex datasets demonstrate that using token style attributes based embedding instead of a raw visual embedding in LayoutLM model is beneficial. Depending on the dataset, such an embedding yields an improvement of 0.18\% to 2.29\% in weighted F1-score with a decrease of 30.7\% in the final number of trainable parameters of the model, leading to an improvement in both efficiency and effectiveness. 
\\


\end{abstract}

\section{Introduction}
Extracting specific information from complex documents is paramount in many business activities. It is a key but complex process because of the wide range of business documents and templates: contracts, invoices, reports or news articles for instance. Despite the recent advances in the field, this task still remains mostly manual and time consuming in various business processes.

Information Extraction is the associated task that aims to automatically extract specific textual information from complex documents. Literature in the domain mainly focuses on two tasks: information extraction from plain text \cite{b2, b3} or information extraction from scanned documents \cite{b4, b5, b6}. While promising, none are completely suited to the growing task of extracting information from machine-readable documents. Indeed, information extraction from plain text is not adapted to document structure underlying complexity that requires not only text related information but also the original layout information \cite{b1} (table structure, paragraphs, two dimensional position on the page...). On the opposite, the complexity of scanned document processing involves the usage of prior text extraction tools \cite{b7} and doesn't allow efficient word-level annotation \cite{b9} (position of the information to extract is approximated). However, information extraction from native PDF documents represents a large and growing number of real-world use cases. This specific task combines the inherent complexity of document processing with the fact that documents, in this case, can be efficiently parsed without approximation and with all the layout attributes (especially text style and position).

In this paper, we introduce a new way of building a visual token representation for the information extraction task on machine-readable documents (\emph{i.e} native PDFs). Indeed, visual information encoding is a key component of most state-of-the-art methods \cite{b1, b8, b9}. Previous approaches rely on encoding the token image representation from the original PDF using computer vision methods like Faster-RCNN \cite{b10} or U-Net \cite{b11}. Although successful, the addition of a computer-vision based approach involves an overall complexity increase of the pipeline, thus increasing the number of parameters and reducing the scalability.

In the context of native PDFs, using dedicated PDF parsing tools, one can directly extract an interpreted version of token visual attributes in what we can call token style: font, font size, color, \emph{etc}. Our approach is based on replacing the original image embedding with embeddings that rely on these style attributes that integrate by nature the meaningful information that can be parsed from the token image.

We experiment a token style variant of the state-of-the-art LayoutLM model on three real-world datasets of native PDF documents (invoices, trade confirmations and fee schedules). We show that it outperforms the image based LayoutLM variant detailed in their paper \cite{b1}.  

Our main contribution can be separated in three parts:

\begin{itemize}
    \item We propose a new intuitive style attributes based embedding in replacement of the image based embedding which outperforms the original approach;
    \item To the best of our knowledge, we provide the first information extraction benchmark of precisely annotated native PDF documents. It is based on three real-world use cases : invoices, trade confirmations, and fee schedules;
    \item We compare the aggregation of style attributes based embedding by addition or by concatenation.
\end{itemize}

\section{Background} 
\subsection{PDF raw content extraction}
\label{ssec:approach-ocr}

The PDF format is broadly adopted and used for exchanging digital documents. In the context of information extraction, the raw content that can be extracted without approximation varies depending on the way the PDF was created. We can distinguish two type of PDFs: the \textit{Scanned PDFs} and the \textit{Native PDFs}. \\

\subsubsection{Scanned PDF Documents}~\\
\label{sssec:scanned}

Scanned PDF Documents consist on images only. The PDF was generated using a scanner or camera and it lost its digital formatting in the process. In the task of information extraction, one need to get textual information from the document. The conversion from a raw set of textual document images to textual content is called Optical Character Recognition (OCR) \cite{b7}. One of the most popular methods is Tesseract OCR Engine \cite{b26}. An OCR Engine converts the set of original images into a textual document $\mathcal{D}_s$ where :
\begin{equation}
\begin{aligned}
  \mathcal{D}_s = {} & (e_s^i)_{i=1...p} \  \text{with} \ e_s^i = (t^i, g^i)
\end{aligned}
\end{equation}
$p$ is the number of tokens in the document\\
$e^i_s$ is the $i$-th token in $\mathcal{D}_s$\\
$t^i$ is the textual content the $i$-th token\\ 
$g^i$ is the coordinates, width and height of the $i$-th token in the document \\

\subsubsection{Native PDF Documents}~\\
\label{sssec:native}

Native PDF Documents are the direct output of an authoring software. It contains all its original formatting like the text, its associated style attributes or the graphics. Hence, native PDFs can be efficiently parsed into a machine-readable document using tools like PDFMiner\footnote{\url{https://github.com/pdfminer/pdfminer.six}} or Apache PDFBox\footnote{\url{https://github.com/apache/pdfbox}}. 
For example, a native PDF document can be converted into a document $\mathcal{D}_n$ where
\begin{equation}
  \begin{aligned}
    \mathcal{D}_n = (e_n^i)_{i=1...p} \ \text{with} \\ 
     e_n^i = (t^i, g^i, b^i, f^i, f_{size}^i, t_{tab}^i, c^i)
    \end{aligned}
\end{equation}
$p, e_n^i, t^i, g^i$ share the denomination from \ref{sssec:scanned} \\
$b^i$ is a boolean assessing if the $i$-th token is bold \\ 
$f^i$ is the font type the $i$-th token\\
$f_{size}^i$ is the font size of the $i$-th token \\
$t_{tab}^i$ is a boolean assessing if the $i$-th token is in a table or not, it can be obtained by parsing the PDF graphics on simple cases or using more advanced computer vision based solution for the extraction \cite{b27}\cite{b28} \\
$c^i$ is the color of the $i$-th token \\

\subsection{Information Extraction}

Information Extraction is the natural language processing task that aims at extracting structured information from unstructured data. In practice, it is often turned into a token classification problem. Given a set of tokens $\mathcal{D} = (e^i)_{i=1...p}$, the goal is to classify each token $e^i$ into a label $l^i$ from a set of labels $\mathcal{L} = (l^i)_{i=1...p}$.

In order to correctly extract entities made of more than one token, IOB (Inside-Outside-Beginning) tagging \cite{b25} is used.
The words prefixed with \textit{I-} (for "Inside") are inside a chunk associated to an entity. The words not corresponding to any entity are labelled as \textit{O} (for "Other"). The first word associated to an entity is prefixed with \textit{B-} (for "Beginning"). This allows us to correctly identify and separate entities inside a text even if they are next to each other.

In the literature, we can distinguish two main applications. Information extraction from plain text that aims at extracting entities from raw textual sentences \cite{b2, b3} and information extraction from scanned documents \cite{b4, b5, b6}. Depending on the task, the information to efficiently model a problem vary. Indeed, layout information is critical for document understanding models as the unique one-dimensional sequential representation used in the prior approaches loses key information \cite{b13}.

\section{Related Work} 
\label{sec:related-work}
\subsection{Information extraction datasets}

Although the literature is rich and a large number of open source datasets have been released, none of them focus on one of the main challenges of extracting information from native PDF documents annotated at the word level. Indeed, as described above, the existing literature focuses on extraction from plain text documents \cite{b2, b3} and extraction from scanned documents \cite{b4, b5, b6}. However, scanned documents processing is an extremely specific domain where raw input is a single image and extraction from plain text doesn't contain the inherent complexity of document processing (tables, title, structure, etc.). 
In our case we propose to focus the research on native PDF Documents with word-level annotation allowing to purely evaluate the information extraction model on the token classification problem. To the best of our knowledge the only complex dataset based on native PDFs is Kleister datasets \cite{b12}. However, entities in this dataset are labeled without their positions. Therefore, we have to use imprecise heuristics to find the exact positions of these entities inside the document. Moreover, Kleister documents are not as visually rich as other datasets (long contract documents).

\subsection{Visual token representation}

Understanding the layout of documents is an obvious step in the task of understanding documents which has been tackled using dedicated methods. Document layout modeling methods were historically numerous. They were mainly graph-based \cite{b13} or based on computer-vision methods enriched with text information \cite{b14, b15}.
Recently, transformers \cite{b16} and pre-trained language models are starting to be widely used for Natural Language Understanding (NLU) with BERT \cite{b17} and its extensions. These models are very powerful to semantically learn a language and can be used for many downstream tasks. However they are built to deal with plain text. 


The first adaptation for document understanding is LayoutLM \cite{b1} that is pre-trained on millions of documents with the additional token coordinates and image representation (see Section \ref{ssec:layoutlm_desc}). More recently, a second version LayoutLMv2 \cite{b8} enriched the first version by including image information in the pre-training framework and the TILT neural network \cite{b9} used a new encoder-decoder architecture that uses text, positions, and image information. LayoutLMv2 and the TILT architectures have slightly higher performances than LayoutLM in some scanned document benchmarks. However, those methods embed complex image embedding during the pre-training phase (while LayoutLM adds it during the fine-tuning) and hence its associated complexity cannot be soften. LayoutLM provides a state-of-the-art baseline with a large architecture overlapping all the other methods. Hence, we will use LayoutLM \cite{b1} enriched with various image embeddings as our main baseline. This, will allow to underline the relevance of replacing image based embeddings by style based embeddings when available without having to reproduce the proprietary pre-training of other methods.

\subsection{Style attributes usage}

Using style attributes in document processing is not a common practice. Indeed as explained, the information is not always available depending on the file format of the document and dominant research is focusing on using token's image representation. On HTML documents, \cite{b18} used handcrafted features based on font size and color in order to improve the performance on information extraction from HTML documents. These results have been extended to other less complex tasks like heading detection as shown by \cite{b19} work. Here, incorporating style based manually crafted features allows to improve standard classifiers. To the best of our knowledge, our work is the first preliminary work integrating style attributes embedding representation in state-of-the-art pre-trained language model based models for the information extraction task.





\section{Information extraction with visual token representation}
\label{sec:approach}
Our approach is largely based on LayoutLM, so it is important to understand this model first. We will discuss about it in the next section. The authors also proposed to add visual embeddings thanks to a computer vision network. This approach is promising but adds a lot of trainable parameters and does not greatly enhance general performances. We will discuss about this idea in section \ref{ssec:approach-image}. Our work is an alternative to the latter, which offers significant improvements while maintaining a reasonable model size.

\subsection{LayoutLM}
\label{ssec:layoutlm_desc}

LayoutLM \cite{b1} architecture is like a BERT Transformer with a slightly different positional embedding. The authors proposed to change it by adding (element-wise sum) a 2D-embeddings on top of the usual 1D-positional embedding.  In fact, knowing all tokens relative position is a key information for information extraction for documents. In short, the positional embedding $\vec{P}^i_{layoutlm}$ of the $i$-th token in a document $\mathcal{D}$ is given by : 

\begin{equation}
\begin{aligned}
&\vec{P}^i_{layoutlm} = \vec{P}^i + \vec{G}^i \\
&\vec{G}^i = \vec{P}^{i}_{x_{1}} + \vec{P}^{i}_{x_{2}} + \vec{P}^{i}_{y_{1}} + \vec{P}^{i}_{y_{2}} + \vec{P}^{i}_{w} + \vec{P}^{i}_{h}
\end{aligned}
\end{equation}
$\vec{P}^i$ : 1D positional embedding introduced in BERT \\
$\vec{G}^i$ : 2D positional embedding introduced in LayoutLM \\
$(\vec{P}^i_j)_{j \in [x_{1},  x_{2}, y_{1}, y_{2}]} $ : embedding of the bounding box coordinates introduced in section \ref{sssec:scanned} \\
$\vec{P}^i_{w}$ : embedding of the bounding box width \\
$\vec{P}^i_{h}$ : embedding of the bounding box height \\

This allows the transformer to use both 1D and 2D position information. For the training phase, they initialized the network with BERT weights and random weights for the 2D embeddings. The model is pre-trained on 2 objectives : Mask Language Modeling (MLM) where some tokens are masked and the model has to predict them; and Multi-label Document Classification (MDC). Pre-training datasets were large enough (11 million scanned images) to build an efficient pre-trained language model adapted to document understanding. Finally, LayoutLM serves as a multi-purpose network and can be used for many downstream tasks, especially information extraction. This approach greatly improved the state-of-the-art on many document understanding tasks such as information extraction for scanned receipts \cite{b4}, and Document classification \cite{b22}. Pre-trained model and code were made available by authors.\footnote{https://github.com/microsoft/unilm/tree/master/layoutlm}

\subsection{LayoutLM with image embedding}
\label{ssec:approach-image}

\begin{figure}[h!]
 \center
  \includegraphics[scale=0.25]{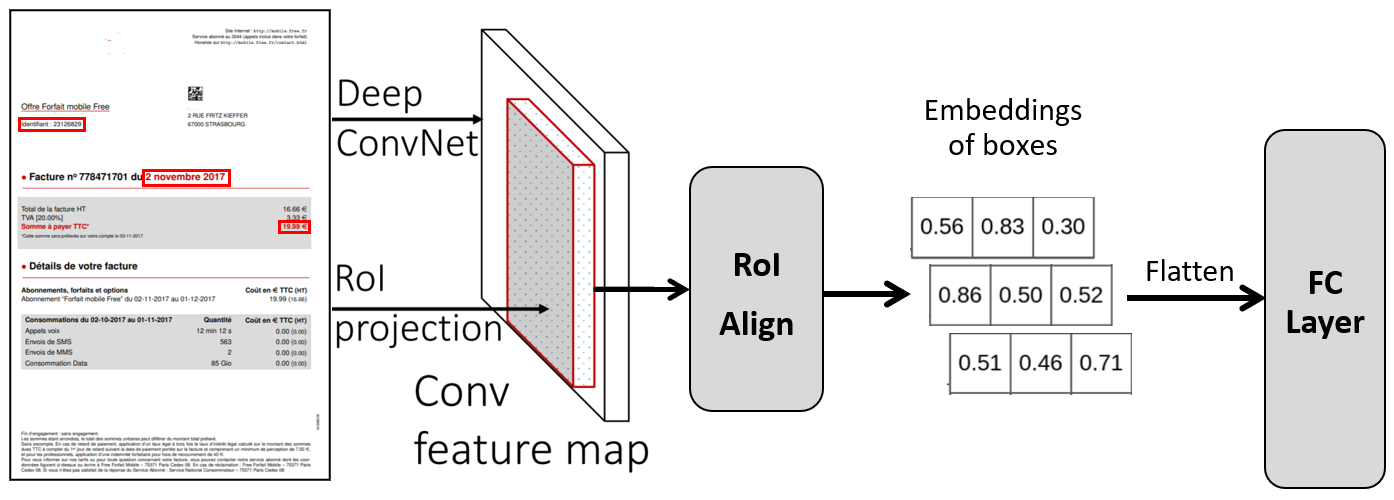}
  \caption{The process of an image embedding}
  \label{process-faster-rcnn}
\end{figure}

The classical LayoutLM model might struggle to understand the full layout of a document as it doesn't depend upon visual information, but only geometric positions. The authors originally propose to add an image embedding following the process illustrated in \autoref{process-faster-rcnn}

They suggest to feed the image $\mathcal{I}$ of a document $\mathcal{D}$ to a Faster R-CNN \cite{b10} that has a backbone $\mathscr{B}$ in order to build a feature map for the document page. The visual feature of a given token should then be extracted from this feature map.

Traditional Regional Proposal Network (RPN) of the Faster R-CNN is not needed as token bounding boxes are given either by an OCR system or by the PDF parsing tool.

Given the document feature map, and token coordinates, the so-called RoIAlign \cite{b21} allows to extract a fixed-size visual representation associated to the token (whatever the token size). This visual embedding can then be projected using linear layers in order to match BERT output dimension.

The overall process can be summarized with the following equation:

\begin{equation}
\label{eq:image-embedding}
\vec{v}_{i} = \phi(\vec{RoIAlign}(\vec{f}^{\mathcal{I}}, g^i))
\end{equation}
$\vec{v}_{i}$ : image embedding of the $i$-th token \\
$\phi$ : linear layer \\
$\vec{f}^{\mathcal{I}}$ : feature map of $\mathcal{I}$ \\
$g^i$ : bounding box of the $i$-th token \\

Token image embedding encompass all the token visual information.
In the original paper \cite{b1}, the authors propose to do an element-wise sum between textual embedding and the visual embedding at the end of the network.

This output token embedding will be fed to a classification layer that will be trained to classify the token in the given categories for the information extraction task. During fine-tuning, all the weights from both LayoutLM and Faster R-CNN are trained.

\subsection{Token Style based Embeddings}

\label{ssec:approach-meta}

Intuitively, image embedding purpose is to capture visual attributes from a token. The current approach involves a tremendous number of trainable parameters and a complexification of models using geometrical and textual information. However, in the case of native PDFs, visual attributes can be directly extracted from the document in what is called styled attributes as explained in \ref{sssec:native}. As an alternative to image embedding, we propose to embed token style attributes instead, as it adds less weights and increases interpretability.

During fine-tuning, we suggest to train a combination of LayoutLM (Section \ref{ssec:layoutlm_desc})
with token style embeddings. As LayoutLM was pre-trained, the additional embedding will be aggregated to LayoutLM original output in order to not corrupt the original pre-training. As detailed previously, the following visual information is available in most PDF parsing tool at token level: 
\begin{itemize}
    \item \emph{bold} : whether the token written in bold,
    \item \emph{inTable} : whether a token belongs to a Table in the document, 
    \item \emph{font} : font type of the token,
    \item \emph{fontSize} : size of the token,
    \item \emph{color} : triplet corresponding to its coordinates in the RGB system.
\end{itemize}

\begin{figure*}
 \center
  \includegraphics[width=\textwidth]{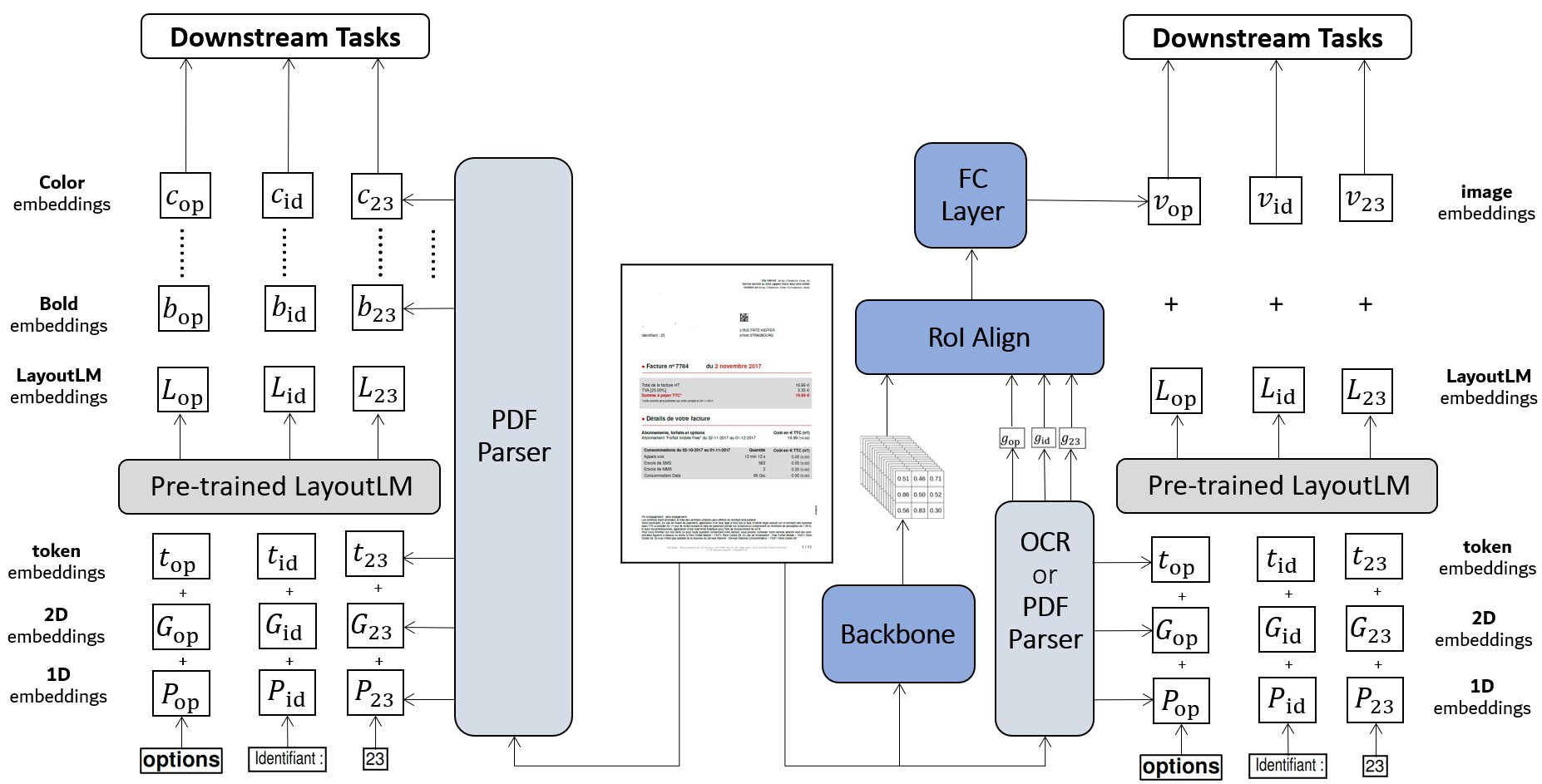}
  \caption{Processes of LayoutLM + style based embeddings (left) and LayoutLM + image embeddings (right)}
  \label{process-layoutlm}
\end{figure*}

Depending on the PDF parser, one could choose what suits best their needs. For example, one can add the style features that assess if a token is in italic or is in a bullet point. Indeed, native PDF format encompass a lot of information that, we think, must be used in the document understanding models.

\autoref{process-layoutlm} shows that the PDF Parser provides the positions and tokens to LayoutLM and extracts meta information that can be used to generate a token style embeddings. Each style feature is linked to an embedding table whose vectors are of size $d_m$. $d_m$ is chosen the same across all style features, but will vary depending on how the fusion with LayoutLM embeddings is done. If we define $d = d_m$, the \textbf{M} embedding tables are of shape ${V_{m}} \times d$ where ${V_{m}}$ is the vocabulary size of the style feature $m$. In the next sections, we will study this number of parameters for two different aggregation methods : element-wise sum or concatenation. 
\\

\subsubsection{Aggregation by concatenation}~\\
\label{ssec:approach-concat}

Concatenating various token embeddings in order to obtain a large and more complete representation of a token is the most traditional way of combining information.
In this specific case, we concatenate the representation vectors for \textbf{M} meta features. 

Concatenating representation involves that the immediate next layer size that projects the embedding into the set of classification labels will be increased. Concatenating allows to be free to chose embedding dimension. For instance, embedding boolean value like \emph{bold} should not require a large dimension. \\

\subsubsection{Aggregation by element-wise sum}~\\
\label{ssec:approach-sum}

Element-wise combination is known to be an efficient way of combining embeddings. Theoretically, it allows the model to learn which embedding feature to prioritize depending on the instance with no prior on which (because all the elements have the same size). 

The dimension of the style features, in this case, is determined by the output dimension of LayoutLM.

\section{Experiments}
\label{sec:xp}
To compare all these methods, we will focus on a key Information Extraction task on three real-world datasets of trade confirmations, invoices, and fee schedules (Section~\ref{ssec:xp-dataset}) which have great complexity. We will compare standard LayoutLM, with LayoutLM model enriched with image embedding, and our proposed method LayoutLM enriched with style based embeddings. \\ In addition, we have also enriched our experiments with tests on two public datasets of scanned documents SROIE \cite{b4} and FUNSD \cite{b5} where only the integration of images can apply as "style" embeddings are not available. \\The experimental protocol is described in detail in Section~\ref{ssec:xp-exp}.

\subsection{Datasets}
\label{ssec:xp-dataset}

\subsubsection{Real-world datasets}

The real-world datasets\footnote{More details in Appendix} are composed of native PDF documents written in multiple languages and containing tables, titles and different font attributes.

They are extremely diverse and commonly shared in the banking business:
\begin{itemize}
    \item Trade Confirmations: 209 one-page PDFs in English with very few original templates (4 font types, 11 font sizes only for instance)
    \item Invoices: 626 one/two-page PDFs written in multiple languages, mostly English and Portuguese, with around 30 different templates. It is diversified (414 font types, 529 font sizes).
    \item FeeSchedule: 273 long PDFs of 1 to 129 English-written pages, with extremely diversified sources (124 font types, 3846 font sizes)
\end{itemize}

They are manually annotated by business experts at word-level, meaning that we know exactly which tokens participate to the entity. 




\subsubsection{Public datasets}

The two public datasets are the most common in the information extraction task. They are based on scanned documents and have been labeled at document level, meaning that we do not know the exact position of the entity in the full document. 
\begin{itemize}
    \item SROIE: 973 scanned receipts in English which were part of ICDAR 2019 competition 
    \item FUNSD: 199 scanned and noisy forms in English
\end{itemize}

\subsection{Experimental protocol}
\label{ssec:xp-exp}

To limit the model size, we chose to group style attributes by hand crafted clusters. 
For example, color embeddings will be grouped into "Black" or "Not Black", font weights are grouped into into 3 clusters (\(\left[0, 1.2\right[\), \(\left[1.2, 2.0\right]\), and \(\left]2.0 +\infty\right[\)). Of course, this processing step is to be adapted to the distribution of the features. 

For the experiments, we used PDFBox in order to parse native PDF documents and extract relevant assets. We used our own implementation of pdftable \cite{b28} combined with the extract graphics assets in order to detect tables. Our LayoutLM model is pre-trained and based on the implementation provided in the transformers library \cite{b23} while the computer vision part is based on torchvision \cite{b24}.

We cross-validated all the models using 5 folds and a batch size of 2 documents. The trainings were made on one NVIDIA Tesla V100 16GB GPU with a constant learning rate: $l_{r}=2\times10^{-5}$ over 20 epochs per iteration of the cross validation. In fact, extending the number of epochs showed no significant improvement. We train with a multi-class cross entropy loss and with Adam optimizer. During the training, we randomly change 10\% of the tokens for improved generalization. We also randomly shift and resize all the bounding boxes to reduce over-fitting during the fine-tuning. In each case, the best model for each iteration of the cross-validation was selected using the validation set.

We used pre-trained LayoutLM base model in order to assess the style embedding relevance. Indeed, one of the main purpose of the proposed work is to limit the number of parameters and LayoutLM large model doesn't outperform the base model by a significant margin.

The main criteria to assess our models is the entity-level weighted average F1 score. F1 scores per model on each dataset (\autoref{table:table1}) is provided for analysis purpose.

Using a grid search over validation set to determine the optimal style embedding dimension, we selected $d=64$ for all the styles when aggregating embeddings using concatenation. Also, when dealing with the aggregation of LayoutLM with style embeddings \begin{savenotes}
    \begin{table*}[hbt!]
      \begin{center}
        \scalebox{1}{
        \begin{tabular}{|c|c|c|c|c|c|c|}
          \cline{3-7}
          \multicolumn{1}{c}{} & \multicolumn{1}{c|}{} & \multicolumn{5}{c|}{5-fold weighted average F1 (\%)} \\
          \hline
          \textbf{Model} & \textbf{Parameters} & \textbf{Trade Confirmations} & \textbf{Invoices} & \textbf{FeeSchedule} & \textbf{SROIE} & \textbf{FUNSD} \\
          \hline
          LayoutLM $+$ ResNet 152        & 163.74M          & 96.81 $\pm$ 2.22                   & 66.29 $\pm$ 2.76 & 75.76 $\pm$ 2.54 & 95.8 $\pm$ 1.26 & 78.42 $\pm$ 2.96\\ 
          LayoutLM-Style Sum           & 113.50M          & 96.76 $\pm$ 0.71                   & 67.60 $\pm$ 2.23 & 76.14 $\pm$ 2.22 & $-$\footnote{Style attribute based approach cannot be used on scanned documents} & $-^{5}$  \\ 
          LayoutLM-Style Concatenation & 113.49M          & \textbf{97.09} $\pm$ \textbf{0.06} & \textbf{68.58} $\pm$ \textbf{1.27} & \textbf{76.64} $\pm$ \textbf{0.01} & $-^{5}$ & $-^{5}$ \\
          \hline
        \end{tabular}
        }
      \end{center}
    \caption{\label{citation-guide} Score \& Number of parameters for LayoutLM, with image embedding, and Style-LayoutLM models}
    \label{table:table1}
    \end{table*}
\end{savenotes} (Style LayoutLM), we use the five features: \emph{bold}, \emph{font}, \emph{fontSize}, \emph{inTable}, and \emph{color} as Style LayoutLM can learn the important style attributes.

To select the backbone used to generate the image embedding, we use a grid search over the set of torchvision available backbones\footnote{https://pytorch.org/vision/stable/models.html}. After multiple experiments, the model studied in this article is a pre-trained ResNet 152.

In order to perform the IOB tagging task, the token embedding output obtained at the end of the pipeline is fed into a dense layer with a dropout rate of 0.3 and softmax as an activation function to predict token classification.

Models are usually limited in terms of sequence length, which is an issue for long documents. For FeeSchedule dataset, we split each document into overlapping shorter documents (called chunks), to aggregate the results at the end. After multiple experiments, we have chosen chunks to be 512 tokens long with an overlap of 100 tokens.

\section{Results}
\label{sec:res}
\subsection{Quantitative results}
\label{ssec:quantitative-results}

In \autoref{table:table1}, we compare the efficiency and the overall performance of the baseline with the models with additional embeddings\footnote{both sum and concatenation style embeddings}.

The image embeddings models have 44.3\% more parameters than LayoutLM. They don't necessarily contribute to enhance the original model, even when the backbone is ResNet 152. Indeed, because of the large volume of parameters, fine-tuning with image embeddings results highly depends on the seed and the fold. It reduces the positive input of LayoutLM pre-training.

The style models increase the number of trainable parameters by 0.01\% as the additional embeddings tables with only thousands of parameters at most. The method allows to integrate visual embedding at a low cost. Even with such a small parameters increase, we could improve the performance in all the datasets for the concatenation fusion. Therefore, style embeddings prove to be a good trade-off between performance and efficiency. 

Regarding the comparison between the models with additional embeddings, the concatenation fusion model happens to be very efficient and consistent in enhancing the overall performance with 30.7\% less parameters than image embeddings model. Also, we get to know which features matter for our task. Independently from the diversity, complexity and length of documents, this aggregation of style based embeddings is the best one as it outperforms LayoutLM in the three datasets in average. 
We also computed a paired t-test on our 5-fold cross validation in our, the associated p-value was 6.89\%. These results show that the approach based on style embedding performs at least as well as the image-based approach with a lower number of parameters, even if the statistical significance of the improvement at the 5\% level is not reached.

However, image-based embeddings takes advantage on public scanned datasets where the fact that style information is not available makes the approach with styles unusable.

\subsection{Qualitative analysis}
\label{ssec:further-analysis}

In the three datasets, there is no significant visual difference between entities of the different classes but the style embedding models happen to improve the detection of some classes\footnote{Scores per class for all datasets available in Appendix}. Regarding the variety of documents, it is difficult for a human to define which attributes will contribute the most in the final model. Random permutations, and training with various combinations of features allows us to define that for the Trade Confirmations dataset the attributes \emph{bold} and \emph{inTable} are the most meaningful while for FeeSchedule and Invoices respectively \emph{fontSize} \& \emph{inTable}, and \emph{inTable} \& \emph{color} are the most importance.

Indeed, looking at the datasets, we can fully understand that:
\begin{itemize}
    \item In the Trade confirmations dataset, entities are often introduced by a word in bold. And many information are presented in a table format,
    \item In Fee Schedules dataset, names of clients are often written with larger fonts and some figures (Rates and Margins) are often presented in tables,
    \item In Invoices datasets, multiple significant figures are described in tables, and some total amount are often in color.
\end{itemize}

\section{Conclusion}
In this paper, we demonstrate the relevance of having dedicated approaches for native PDF documents where token style attributes can be extracted. We confirm the intuitive result that having the style attributes instead of the raw image in order to provide visual information to the model we were able to obtain better performances in the information extraction task while reducing the number of parameters. While benchmarked against state-of-the-art LayoutLM  we think that this work can be applied to other approaches and could be extended by using font style embeddings during the pre-training process.


\newpage
\appendices

\setcounter{section}{0}

\section{Datasets}
\label{sec:datasets}
\subsection{Trade Confirmations}

\label{ssec:datasets-derivatives}

Trade confirmations are financial documents that summarise all the details of one of multiple financial trades. It is usually well structured with tables clear structure. Our dataset consist of derivatives products with the following business related labels:

\begin{itemize}
    \item BUYSELL : direction of the trade (buy or sell)
    \item CALLPUT : is it a call or a put (the contract allows to buy or sell the option)
    \item CLEAR INFO : account number
    \item CONTRACT : contract number
    \item EXECUTIVE BROKER : broker associated to the trade
    \item EXPIRY DATE : maturity of the trade
    \item MARKET : market where the trade applies
    \item TRADE PRICE : contract price
    \item STRIKE VALUE: fixed price to buy or sell the underlying as defined in the contract 
    \item TRADE DATE : date of the trade
    \item TRADE STATUS : status of the trade
    \item TRADE VOLUME : quantity to buy 
\end{itemize}

BUYSELL and CALLPUT are easier to extract because the majority of datapoints take their values in a finite set of words (buy, sell, call, put ...). On the contrary TRADE PRICE and TRADE VOLUME are more difficult to parse because they are often displayed within multiple tables and require layout understanding in order to be efficiently extracted.

\subsection{Invoices}
\label{ssec:datasets-invoices}

Invoices are the classic document that shows what goods were purchased and at what price. Our annotators labeled the following entities:
\begin{itemize}
    \item ACCOUNTNUMBER : account ID
    \item ADDRESS : all the addresses in the document
    \item COMPANYNAME : company name
    \item DOCUMENTDATE : date of the document
    \item GROSSWEIGHT : the weight of an item (optional)
    \item IBAN : relates to the account paying the invoice
    \item INVOICENUMBER : invoice ID
    \item PERSONNAME : name of a person
    \item TOTAL : total amount to be paid
\end{itemize}

As described above, the dataset difficulty comes from the wide range of templates within the documents. \\
Moreover, some datapoints are particularly hard to extract: ADDRESS are difficult to parse because it is made up from 5 to 10 words written in multiple lines, COMPANYNAME is made up of non common words and often comes without easy to parse context, finally GROSSWEIGHT is often displayed in tables without border that requires the model to gain a complex layout understanding.

\subsection{FeeSchedule}
\label{ssec:datasets-feeschedule}

FeeSchedule is dataset of financial documents that detail a pricing proposal for a specific client. The most important information are the interest rates applied to the client. They are composed of two parts: a base rate and a margin. 
\\
The labeled entities are :
\begin{itemize}
    \item APPLICATION DATE : date of the pricing
    \item BRANCH NAME : the branch of the bank responsible for the proposal
    \item CLIENT NAME : name of the concerned client
    \item MARGIN : base rate of the Fees
    \item RATE : the margin associated to the rate
\end{itemize}

The dataset consist of documents where the fees are presented in tables of different structures, but also inside the paragraphs in some cases. \\
All FeeSchedule documents are passed through the OCR before the processing which resulted in errors in the retrieved text especially for the data of the cover page and the signature page. \\ 
This affects mainly the datapoints APPLICATION DATE, BRANCH NAME and CLIENT NAME, which were often tagged with inconsistent values. Indeed, sometimes information can be handwritten or can be presented in multiple parts of the document which makes labelling consistently hard and in consequence impacts the overall performances for those datapoints. 

\section{Detailed performances}
\label{sec:performances}

\begin{table*}[hbt!]
  \centering
  \scalebox{1.3}{
  \begin{tabular}{c|c c c c}
     \textbf{Label} & \textbf{Image Embeddings} & \multicolumn{2}{c}{\textbf{Style Embeddings}} & \textbf{Support} \\
    \cline{1-5}
    \textbf{} & \textbf{R152} & \textbf{Concatenation} & \textbf{Sum} \\
    \hline
    BUYSELL          & 99.34 & 99.15 & 98.98 & 236 \\ \hline
    CALLPUT          & 96.13 & 96.68 & 93.13 & 117 \\ \hline
    CLEAR INFO       & 97.29 & 96.83 & 97.05 & 207 \\ \hline
    CONTRACT         & 96.88 & 97.80 & 97.33 & 207 \\ \hline
    EXECUTIVE BROKER & 97.37 & 97.33 & 96.83 & 208 \\ \hline
    EXPIRY DATE      & 97.20 & 96.80 & 97.84 & 229 \\ \hline
    MARKET           & 98.31 & 98.58 & 98.30 & 207 \\ \hline
    TRADE PRICE            & 88.32 & 89.93 & 90.83 & 237 \\ \hline
    STRIKE VALUE     & 99.78 & 99.59 & 99.37 & 231 \\ \hline
    TRADE DATE       & 100   & 99.52 & 99.52 & 207 \\ \hline
    TRADE STATUS     & 100   & 100   & 99.68 & 161 \\ \hline
    TRADE VOLUME           & 92.56 & 94.22 & 92.22 & 238 \\ \hline
  \end{tabular}
  }
  \caption{\label{citation-guide} Average 5-fold F1 score (\%) per class for all models on Trade Confirmations}
  \label{table:table1}

\end{table*}

\begin{table*}[hbt!]
  \centering
  \scalebox{1.3}{
  \begin{tabular}{c|c c c c}
     \textbf{Label} & \textbf{Image Embeddings} & \multicolumn{2}{c}{\textbf{Style Embeddings}} & \textbf{Support} \\
    \cline{1-5}
    \textbf{} & \textbf{R152} & \textbf{Concatenation} & \textbf{Sum} \\
    \hline
    ACCOUNTNUMBER & 73.78 & 75.53 & 75.96 & 463 \\ \hline
    ADDRESS       & 57.26 & 61.46 & 59.46 & 2037 \\ \hline
    COMPANYNAME   & 64.24 & 65.65 & 64.68 & 2164 \\ \hline
    DOCUMENTDATE  & 77.89 & 76.67 & 78.86 & 555 \\ \hline
    GROSSWEIGHT   & 45.47 & 49.85 & 43.93 & 244 \\ \hline
    IBAN          & 86.60 & 90.76 & 86.43 & 258 \\ \hline
    INVOICENUMBER & 74.25 & 75.73 & 74.32 & 581 \\ \hline
    PERSONNAME    & 74.35 & 75.80 & 76.28 & 1096 \\ \hline
    TOTAL         & 65.84 & 69.59 & 70.40 & 538 \\ \hline

  \end{tabular}
  }
  \caption{\label{citation-guide} Average 5-fold F1 score (\%) per class for all models on Invoices}
  \label{table:table2}

\end{table*}

\begin{table*}[hbt!]
  \centering
  \scalebox{1.3}{
  \begin{tabular}{c|c c c c}
     \textbf{Label} & \textbf{Image Embeddings}  & \multicolumn{2}{c}{\textbf{Style Embeddings}} & \textbf{Support} \\
    \cline{1-5}
    \textbf{} & \textbf{R152} & \textbf{Concatenation} & \textbf{Sum} \\
    \hline
    APPLICATION DATE & 41.74 & 42.90 & 32.07 & 273 \\ \hline
    BRANCH NAME      & 18.27 & 23.22 & 27.67 & 270 \\ \hline
    CLIENT NAME      & 23.40 & 29.19 & 20.61 & 274 \\ \hline
    MARGIN           & 81.32 & 81.44 & 81.44 & 3907 \\ \hline
    RATE             & 78.71 & 79.67 & 79.46 & 5913 \\ \hline

  \end{tabular}
  }
  \caption{\label{citation-guide} Average 5-fold F1 score (\%) per class for all models on FeeSchedule}
  \label{table:table3}

\end{table*}

\end{document}